\documentclass{article}
\usepackage{spconf,amsmath,graphicx,hyperref}
\usepackage{amssymb,amsfonts}
\usepackage{multirow}
\usepackage{multirow,multicol}
\usepackage{algorithmic}
\usepackage{graphicx}
\usepackage{booktabs}
\usepackage{textcomp}
\usepackage{xcolor}
\usepackage{rotating}
\usepackage[numbers,sort&compress]{natbib}
\title{DSVM-UNet : Enhancing VM-UNet with Dual Self-distillation for Medical Image Segmentation}
\name{Renrong Shao$^{1}$, Dongyang Li$^{2}$, Dong Xia$^{1}$, Lin Shao$^{1}$, Jiangdong Lu$^{1}$, Fen Zheng$^{1}$, Lulu Zhang$^{1}$}
\address{$^{1}$Faculty of Military Health Service, Naval Medical University, China\\
$^{2}$School of Computer Science and Technology, Shanghai University of Electric Power, China\\
}
\begin{document}
\maketitle
\begin{abstract}
Vision Mamba models have been extensively researched in various fields, which address the limitations of previous models by effectively managing long-range dependencies with a linear-time overhead. 
Several prospective studies have further designed Vision Mamba based on UNet~(VM-UNet) for medical image segmentation. 
These approaches primarily focus on optimizing architectural designs by creating more complex structures to enhance the model's ability to perceive semantic features. 
In this paper, we propose a simple yet effective approach to improve the model by \textbf{D}ual \textbf{S}elf-distillation for \textbf{VM-UNet}~(\textbf{DSVM-UNet}) without any complex architectural designs. To achieve this goal, we develop double self-distillation methods to align the features at both the global and local levels.
Extensive experiments conducted on the ISIC2017, ISIC2018, and Synapse benchmarks demonstrate that our approach achieves state-of-the-art performance while maintaining computational efficiency. Code is available \href{https://github.com/RoryShao/DSVM-UNet.git}{\textit{here}}.
\end{abstract}
\section{Introduction}
\label{sec:intro}
Semantic segmentation, as a fundamental task in medical image analysis, has been receiving widespread attention. Recent advancements in artificial intelligence, especially deep learning, have led to remarkable success in this area.
Currently, the U-shaped encoder-decoder architecture with integral skip connections for semantic segmentation~\cite{vnet, shao2018general, zhou2018unet++, transunet, swinunet, ruan2024vm, chen2024msvm}, represented by UNet~\cite{unet}, has become the de facto standard model. The model can effectively preserve key spatial information and merge features from the encoder and decoder layers for addressing the segmentation issues of complex structures.
The model's success stems from its ability to retain key spatial information efficiently while merging features from the encoder and decoder layers to address the segmentation challenges of complex structures.
As medical research advances and the complexity of medical images increases, UNet variants relying solely on conventional CNNs can no longer effectively capture long-distance dependencies. This limitation hinders segmentation accuracy in complex scenes. 
While variants relying solely on Transformer~\cite{vaswani2017attention,liu2021swin,shao2025consistent} effectively address long-range dependency issues, the self-attention mechanism involves complex matrix computations that can lead to significant computational overhead and increased processing costs.
Recent advancements have introduced the innovative architecture known as Mamba~\cite{gu2023mamba}, which is highly effective in efficiently capturing global contextual information. Mamba is specifically designed for long-range modeling and is recognized for its computational efficiency based on state space models (SSM)~\cite{ssm-s4}. Following this, Vision Mamba~\cite{vim2024zhu} and VMamba~\cite{liu2024vmamba} have expanded Mamba's architecture into the realm of computer vision, improving upon Mamba's unidirectional scanning mechanism. Given its efficient ability to model global context, the Mamba architecture is highly suitable for processing medical image segmentation. 
Currently, several prospective studies combine Vision Mamba with UNet networks~(VM-UNet) to achieve image segmentation in the medical field. 
These approaches primarily focus on optimizing architectural designs to enhance the model's ability to perceive semantic features. 
\begin{figure*}[!h]
    \vspace{-2mm}
    \centering
    \includegraphics[width=0.90\textwidth]{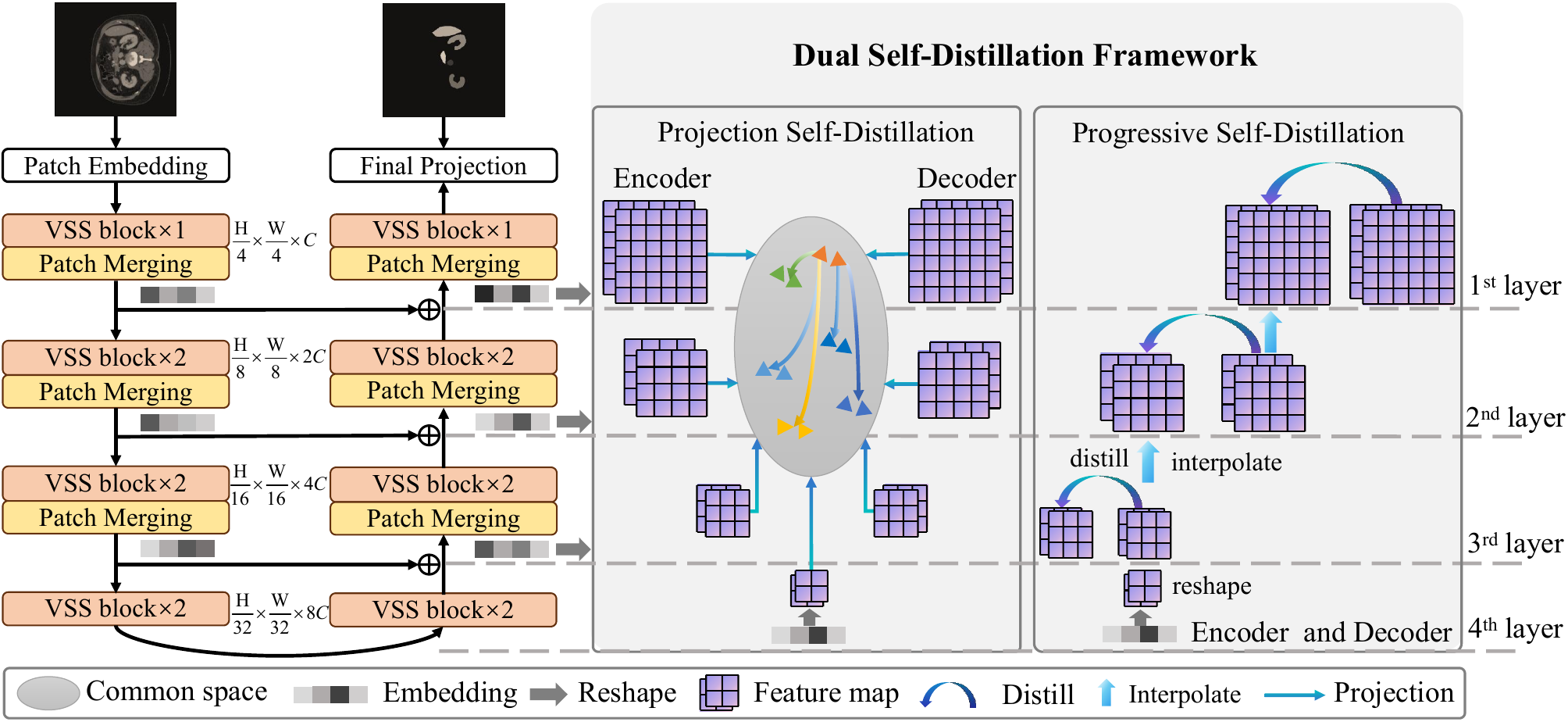}
    \vspace{-2mm}
    \caption{The entire framework of our DSVM-UNET methods is illustrated. On the right is the dual self-distillation framework, which comprises projection self-distillation and progressive self-distillation. VSS blocks are derived from VM-UNet~\cite{ruan2024vm}.}
    \label{fig:framework}
    \vspace{-4mm}
\end{figure*}

In this paper, we propose a simple yet effective approach to improve the model by \textbf{D}ual \textbf{S}elf-distillation for \textbf{VM-UNet}~(\textbf{DSVM-UNet}) without any complex architectural designs.
This method first obtains feature representations from the intermediate layers of the UNet-shaped Vision Mamba and aligns these representations at both the global and local levels. 
Specifically, dual self-distillation methods, such as progressive self-distillation and projection self-distillation, facilitate the transfer of knowledge from high-level features to low-level features in the network, enabling the spatial alignment of various semantic features.
Without any other flashy architectural designs, our method solely achieves performance improvement through the mining of semantic features and distillation.
Extensive experiments conducted on the ISIC2017, ISIC2018, and Synapse benchmarks demonstrate that our approach achieves state-of-the-art~(SOTA) performance while maintaining computational efficiency.

In a nutshell, our contributions are three-fold: (1) We propose DSVM-UNet, a simple yet effective UNet-style segmentation dual self-distillation framework that can facilitate knowledge from high-level features to low-level features in the network. (2) We propose two self-distillation approaches, such as projection self-distillation and progressive self-distillation, which can effectively eliminate global and local variations of features to improve the medical image segmentation performance. (3) We conduct extensive experiments on multiple benchmarks, demonstrating that DSVM-UNet achieves SOTA performance while remaining computationally efficient.

\section{Methodology}
\label{sec2}
\subsection{Preliminaries}

\textbf{State Space Models.} Modern SSM-based models, such as S4 and Mamba~\cite{liu2024vmamba,zhang2024vm}, leverage a powerful classical continuous system. This innovative approach effectively transforms a one-dimensional input function or sequence, represented as $ x(t) \in \mathcal{R} $, through a series of intermediate implicit states $ h(t) \in \mathcal{R}^N $, ultimately yielding a precise output $y(t) \in \mathcal{R}$. This entire transformation is captured by a linear ordinary differential equation (ODE):
\begin{equation}
  \small
\begin{aligned}
    &h^{\prime}(t) = \mathbf{A}h(t) + \mathbf{B}x(t) \\
    &y(t) = \mathbf{C}h(t)
\end{aligned}
\end{equation} where $\mathbf{A} \in \mathcal{R}^{N \times N}$ represents the state matrix, while $\mathbf{B} \in  \mathcal{R}^{N \times 1}$ and $\mathbf{C} \in  \mathcal{R}^{N \times 1}$ denote the projection parameters.

\noindent\textbf{Vision Mamba Unet.} VM-UNet comprises a Patch Embedding layer, an encoder, a decoder, a Final Projection layer, and skip connections. 
The core module in the encoder and decoder of VM-UNet is the VSS black, which is derived from VMamba~\cite{liu2024vmamba}.
Given an input image $x$, where $ x \in \mathcal{R}^{\text{H} \times \text{W}\times\text{3}}$, the encoder and decoder generate features at $M$ levels. Each level features can be denoted as $f_{l} \in \mathcal{R}^{2^{l-1}\text{C}\times \frac{\text{H}}{2^{l+1}}\times\frac{\text{W}}{2^{l+1}}}$, where $1 \leq l\leq M $ and $C$ represents dimensions of non-overlapping patches mapped by Patch Embedding layer.  

\subsection{Dual Self-distillation Framework}
Currently, prospective studies~\cite{ruan2024vm,zhang2024vm,lei2024vm,liao2024lightm} mainly focus on the design of the model architecture to enhance the ability of perceiving semantic features.
Unlike existing approaches, our method primarily operates at the multi-level features and mines semantic features to address knowledge gaps through self-distillation.
Therefore, to improve the model's performance without increasing its overhead, we propose the DSVM-UNet with double distillation methods, i.e., projection self-distillation and progressive self-distillation.

\vspace{-4mm}
\subsubsection{Projection Self-Distillation}
DSVM-UNet utilizes a standard encoder-decoder architecture as~\cite{ruan2024vm,zhang2024vm}, as illustrated in Fig~\ref{fig:framework}. 
Due to its architecture, the model can extract features at multiple scales,  resulting in semantic variation among the features in the intermediate layers.
To eliminate this inconsistency, we design the projection distillation method. It aligns encoder and decoder features simultaneously by projecting them into a common space and measuring their differences.
We first obtain each feature embeddings from the intermediate layers and reshape them as $f_{l} \in \mathcal{R}^{2^{l-1}\text{C}\times \frac{\text{H}}{2^{l+1}}\times\frac{\text{W}}{2^{l+1}}}$, 
where $l$ represents the corresponding layer and subject to $l\in [1, M]$, $M$ is the total layers of encoder or decoder.
Due to the inconsistency in scale of these features, making direct alignment is impossible. 
Therefore, we need to align them in both the channel-wise and spatial-wise.
Specifically, we first align the spatial-wise features by linear projection as $\text{Lin}(f_{l})\in \mathcal{R}^{2^{l-1}\text{C}\times\frac{\text{H}}{4}\times\frac{\text{W}}{4}}$, which projects all level features into the same space.
Next, we project all level features by $1\times1$ 1D convolution $\text{Conv}(\text{Lin}(f_{l}))$ to obtain aligned channel-wise features as $\hat{f}_{l}\in \mathcal{R}^{\text{C}\times\frac{\text{H}}{4}\times\frac{\text{W}}{4}}$.  
To further eliminate the global difference of different levels, we utilize the last layer feature $f_{1}$ of the decoder as the teacher due to its high-level feature representations. 
Then, align all level features to those of the final layer, respectively.
Finally,  the distillation loss is constructed on the features as follows Eq.~\ref{eq:proj_distill}:
\begin{equation}
  \vspace{-1mm}
  \small
    \begin{aligned}
      \label{eq:proj_distill}
   & \hat{f} _{l} = \mathrm{Conv1D}(\mathrm{Lin}(f_{l})) \,, \\ 
   & \mathcal{L}_{\text{Proj}} = \sum^{M}_{l=1} \mathrm{Distill}(\hat{f}^{e}_{l}, {f}^{d}_{1}) + \sum^{M-1}_{l=1} \mathrm{Distill}(\hat{f}^{d}_{l}, {f}^{d}_{1}) \,,
    \end{aligned}
    \vspace{-1mm}
\end{equation}
where $\hat{f} _{l}$ denotes the projected feature integrating all level features, while $\hat{f}^{e}_{l}$ and $\hat{f}^{d}_{l}$ represent the encoder and decoder features, respectively. Lin($\cdot$) signifies the linear projection, and Distill($\cdot$) denotes the distillation operation, which is measured by the mean squared error (MSE) loss.
$\mathcal{L}_{\text{Proj}}$ ensures that each layer learns high-level semantic features, thereby compensating for the limited learning capacity of shallow models.

\subsubsection{Progressive Self-Distillation}
To further eliminate the local difference of different levels, we design the progressive self-distillation, which consists of two steps, i.e., encoder distillation and decoder distillation. 
Both of them adhere to the principle that when aligning features between adjacent layers, features with lower dimensionality are always aligned to those with higher dimensionality.
For example, when exploiting encoder distillation, we first obtains the all level features from each layer of the intermediate layers as $\{f^{e}_{1}, f^{e}_{2}, f^{e}_{3}, f^{e}_{4}\}$, in which $f^{e}_{1}\in \mathcal{R}^{\text{C}\times\frac{\text{H}}{4}\times\frac{\text{W}}{4}}$ and  $f^{e}_{2}\in \mathcal{R}^{\text{2C}\times\frac{\text{H}}{8}\times\frac{\text{W}}{8}}$. 
Then, to ensure that the features of two layers have consistent dimensional representations, we apply 1$\times$1 2D convolution to $f^{e}_{2}$ to achieve consistent channels with $f^{e}_{1}$, i.e.,  $\text{Conv2D}(f^{e}_{2}) \in \mathcal{R}^{\text{C}\times\frac{\text{H}}{8}\times\frac{\text{W}}{8}}$.
To further achieve spatial dimensions alignment of two layers, we then exploit bilinear interpolation to scale up the dimension of $f^{e}_{2}$ as $\tilde{f}^{e}_{1} = \text{Inp}(\text{Conv2D}(f^{e}_{2}))\in\mathcal{R}^{\text{C}\times\frac{\text{H}}{4}\times\frac{\text{W}}{4}}$. 
Likewise, the decoder also achieves all level features from each layer of the intermediate as $\{f^{d}_{4}, f^{d}_{3}, f^{d}_{2}, f^{d}_{1}\}$. 
For $ f^{d}_{2}$ and $f^{d}_{1}$, we need to adopt a dual operation similar to that of an encoder, i.e., convolution is first employed to achieve consistency between $f^{d}_{2}$ and $f^{d}_{1}$ in the channel-wise features, followed by interpolation to ensure consistency in the spatial-wise features.
Given that the latter layer of the whole model contains a higher-level semantic feature than the former features of adjacent layers, we identify the latter feature as the teacher.
Finally, we progressively distill the knowledge between adjacent feature layers as follows:
\begin{equation}
  \small
    \begin{aligned}
      \label{eq:prog_distill}
   & \tilde{f}_{l} = \mathrm{Inp}(\mathrm{Conv2D}(f_{l}))  \,,  \\
   & \mathcal{L}_{\text{Prog}} = \sum^{M}_{l=2} \left[ \mathrm{Distill}(\tilde{f}^{e}_{l-1}, \tilde{f}^{e}_{l}) + \mathrm{Distill}(\tilde{f}^{d}_{l}, \tilde{f}^{d}_{l-1})\right]  \,,
    \end{aligned}
\end{equation}
where $\tilde{f}_{l}$ represents the aligned features of the encoder and decoder, Inp($\cdot$) stands for the bilinear interpolate operation, the distillation operation Distill($\cdot$) here, exploiting L1 loss, is mainly used to be performed for two adjacent block features.

\subsection{Total Loss}
Besides, our approach is built upon fundamental loss. For binary segmentation, we use the Binary Cross-Entropy and Dice loss (BceDice loss, Eq.~\ref{eq:bcediceloss}) with $\mathcal{L}_{\text{Proj}}$ and $\mathcal{L}_{\text{Prog}}$ for total loss Eq~\ref{eq:total_loss}. 
For multi-class segmentation, we replace $\mathcal{L}_{\text{BceDice}}$ in Eq.~\ref{eq:total_loss} with the Cross-Entropy and Dice loss, following \cite{zhang2024vm}.
\begin{equation}
  \small
   \begin{aligned}
    & \mathcal{L}_{\text{BceDice}} = \lambda_1 \cdot \mathcal{L}_{\text{Bce}} + \lambda_2 \cdot \mathcal{L}_{\text{Dice}} \,, \\ 
    & \mathcal{L}_{\text{Bce}} = -\frac{1}{N} \sum_{i=1}^{N} \left[ y_i \log(\hat{y}_i) + (1 - y_i) \log(1 - \hat{y}_i) \right] \,, \\
    & \mathcal{L}_{\text{Dice}} = 1 - \frac{2|X \cap Y|}{|X|+|Y|} \,.
   \end{aligned}
    \label{eq:bcediceloss}
\end{equation} 
Finally, the total loss is obtained as follows Eq.~\ref{eq:total_loss}:
\begin{equation}
  \small
    \mathcal{L}_{\text{Total}} = \mathcal{L}_{\text{BceDice}}+ \alpha\mathcal{L}_{\text{Proj}} + \beta \mathcal{L}_{\text{Prog}}  \,,
    \label{eq:total_loss}
\end{equation}
where $\mathcal{L}_{\text{BceDice}}$ served as the baseline for our experiments, $\alpha$, $\beta$ are hyperparameters to balance the loss.

\begin{table}[!t]
  \centering
	\caption{Comparative experimental results on the ISIC17 and ISIC18 dataset. (\textbf{Bold} indicates the best.)}
  \vspace{1mm}
    \resizebox{\linewidth}{!}{
		\begin{tabular}{c|c|ccccc}
      \toprule
      \textbf{Dataset} &\textbf{Model}  & \textbf{mIoU(\%)$\uparrow$}  & \textbf{DSC(\%)$\uparrow$}  &  \textbf{Acc(\%)$\uparrow$}   & \textbf{Spe(\%)$\uparrow$}   & \textbf{Sen(\%)$\uparrow$}   \\ 
      \midrule
      \multirow{10}{*}{ISIC17} &UNet \cite{unet} &76.98 &86.99 &95.65 &97.43 & 86.82  \\
      &UTNetV2 \cite{utnetv2} & 77.35  & 87.23  & 95.84  & 98.05  & 84.85  \\
      &TransFuse \cite{transfuse} & 79.21  & 88.40  & 96.17  & 97.98  & 87.14 \\
      &MALUNet \cite{malunet} &78.78 &88.13 &96.18 &\textbf{98.47} &84.78 \\
      &VM-UNet \cite{ruan2024vm} & 80.23 & 89.03  & 96.29  & 97.58 & 89.90  \\  
      &VM-UNet++ \cite{lei2024vm}  & 80.49 & 89.19 & 96.44 & 98.24 & 87.53 \\
      &ASP-VMUNet \cite{bao2025asp} & 75.63 & 86.12 & 93.77 & 97.44 & 81.89 \\
      &VM-UNetV2 \cite{zhang2024vm} &  82.34 & 90.31 & \textbf{96.70} & 97.67 & 91.89   \\ 
      & DSVM-UNet~(\textbf{Ours}) & \textbf{82.57}  & \textbf{90.62} & 96.69 & 98.34 & \textbf{92.08} \\
      \midrule
      \multirow{13}{*}{ISIC18}&UNet \cite{unet} & 77.86  & 87.55  & 94.05  & 96.69  & 85.86  \\
      &UNet++ \cite{zhou2018unet++} & 78.31  & 87.83  & 94.02  & 95.75  & 88.65  \\
      &Att-UNet \cite{attentionunet}  & 78.43  & 87.91  & 94.13  & 96.23  & 87.60  \\
      &TransFuse \cite{transfuse} & 80.63  & 89.27  & 94.66  & 95.74  & \textbf{91.28} \\
      &MALUNet \cite{malunet} &80.25 &89.04 &94.62 &96.19 &89.74 \\
      &VM-UNet \cite{ruan2024vm} & 81.35 & 89.71 & 94.91  &96.13 &91.12   \\ 
      &VM-UNet++ \cite{lei2024vm} &  80.17 & 88.99  & 94.67  &  96.64 & 88.52 \\
& ASP-VMUNet ~\cite{bao2025asp} & 80.03 & 88.91 & 93.84 & 96.02 & 88.23 \\
      &VM-UNetV2  \cite{zhang2024vm} & 81.37 & 89.73 & 95.06 & 97.13 & 88.64 \\
      & DSVM-UNet~(\textbf{Ours}) & \textbf{81.51} & \textbf{90.45} &\textbf{95.43} & \textbf{97.66} & 91.26 \\
    \bottomrule
		 \end{tabular}
     }
		\label{tab:isic}
    \vspace{-4mm}
\end{table}
  
\begin{table*}[!t]
  \centering
	\small
	\caption{Comparative experimental results on the Synapse dataset. DSC of every single class (Aorta, Gallbladder, Kidney(L), Kidney(R), Liver, Pancreas, Spleen, Stomach) is also reported. (\textbf{Bold} indicates the best.)}
  \vspace{1mm}
  \resizebox{0.78\linewidth}{!}{
		\begin{tabular}{c|cc|cccccccc}
      \toprule
			\textbf{Model}  & \textbf{DSC$\uparrow$} & \textbf{HD95$\downarrow$} & \textbf{Aor.} & \textbf{Gal.} & \textbf{Kid.(L)} & \textbf{Kid.(R)} & \textbf{Liv.} & \textbf{Pan.} & \textbf{Spl.} & \textbf{Sto.} \\ 
			\midrule
    DARR \cite{DARR}  &69.77 &- &74.74 &53.77 &72.31 &73.24 &94.08 &54.18 &89.90 &45.96 \\
    R50 U-Net \cite{transunet} &74.68 &36.87 &87.47 &66.36 &80.60 &78.19 &93.74 &56.90 &85.87 &74.16 \\
			UNet \cite{unet}     & 76.85    & 39.70     & 89.07 & 69.72  & 77.77 & 68.60 & 93.43  & 53.98     & 86.67   & 75.58    \\
		Att-UNet \cite{attentionunet}   & 77.77    & 36.02     & \textbf{89.55}  & 68.88   & 77.98 & 71.11 & 93.57  & 58.04     & 87.30   & 75.75    \\
		TransUnet \cite{transunet} & 77.48    & 31.69     & 87.23  & 63.13   & 81.87 & 77.02 & 94.08   & 55.86     & 85.08   & 75.62    \\
    TransNorm \cite{transnorm}  &78.40 &30.25 &86.23 &65.10 &82.18 &78.63 &94.22 &55.34 &89.50 &76.01\\
    Swin U-Net \cite{swinunet}  &79.13 &21.55 &85.47 &66.53 &83.28 &79.61 & 94.29 &56.58 & 90.66 &76.60 \\
    TransDeepLab \cite{transdeeplab} &80.16 &21.25 &86.04 &69.16 &84.08 &79.88 &93.53 &\textbf{61.19} &89.00 &78.40 \\
    MT-UNet  \cite{mtunet}  & 78.59    & 26.59     & 87.92  & 64.99   & 81.47 & 77.29 & 93.06  & 59.46    & 87.75   & 76.81   \\ 
	  MEW-UNet \cite{mewunet} & 78.92   & \textbf{16.44}    & 86.68  & 65.32   & 82.87     & 80.02 & 93.63  & 58.36     & 90.19  & 74.26  \\ 
    VM-UNet \cite{ruan2024vm} &81.08 &19.21&86.40&69.41& 86.16 & 82.76&94.17&58.80&89.51& 81.40  \\
    VM-UNetV2 \cite{lei2024vm} & 81.21 & 20.64 & 86.38 & 69.61 & \textbf{86.65} & 82.93 & 94.34 & 58.91 & 89.66 & 81.59 \\
  \midrule
    DSVM-UNet~\textbf{(Ours)} & \textbf{81.68}  & 19.32 & 87.56 & \textbf{69.90} & 86.61 & \textbf{83.32} & \textbf{94.44} & 59.25 & \textbf{90.67} & \textbf{81.72} \\
  \bottomrule
		\end{tabular}
  }
		\label{tab:synapse}
  \vspace{-4mm}
\end{table*}

\section{experiments}
\subsection{Implementation Details}
Following the prior works~\cite{ruan2024vm,zhang2024vm}, we adjust the image dimensions in all datasets
to 256$\times$256. To mitigate overfitting, data augmentation techniques such as random flipping and random rotation are exploited. For the ISIC17 and ISIC18 datasets, the BceDice loss function is utilized, while the CeDice loss function is applied to the Synapse dataset. The batch size is set to 32, and we use the AdamW \cite{adamw} optimizer with an initial learning rate of 1e-3. The learning rate is scheduled using CosineAnnealingLR \cite{cosineannealingLR}, with a maximum of 50 iterations and a minimum learning rate of 1e-5. The training epochs are set to 300, with the parameters $\alpha$ and $\beta$ assigned values of 1 and 0.5, respectively. For the DSVM-UNet, we initialize the weights of both the encoder and decoder with those from VMamba-S \cite{liu2024vmamba}, which was pre-trained on ImageNet-1k. All experiments are conducted on NVIDIA RTX A40 GPUs.

\subsection{Comparison to the SOTA Models}
\textbf{ISIC Datasets.} We compare our DSVM-UNet with many SOTA models on the ISIC17 and ISIC18 datasets. In Table \ref{tab:isic}, VM-UNet series demonstrate strong performance in skin lesion segmentation tasks. Specifically, on the ISIC17 dataset, VM-UNetV2 achieves the best results in four metrics: mIoU, DSC, Acc, and Sen. For the ISIC18 dataset, VM-UNetV2 also outperforms other models in four metrics: mIoU, DSC, Acc, and Spe. 
Particularly, compared to VM-UNetV2 on the ISIC17 dataset, a powerful enhanced VM-UNet model, our DSVM-UNet achieves a 0.23\% increase in mIoU, a 0.31\% increase in DSC, a 0.67\% increase in Spe, and a 0.19\% increase in Sen. Similar improvements are observed on the ISIC18 dataset as well.

\noindent\textbf{Synapse Dataset.} 
For the Synapse dataset, we conduct a comprehensive comparison with existing SOTA models, as presented in Table \ref{tab:synapse}. 
These models can be divided into CNN-based models~\cite{transunet,unet,attentionunet,DARR}, Transformer-based models~\cite{transunet,swinunet,transnorm}, hybrid models~\cite{transdeeplab,mtunet,mewunet}, and VM-Unet models~\cite{ruan2024vm,lei2024vm}. Both of them are dependent on architectural improvements.
Our approach primarily utilizes self-distillation to address issues at the feature level of the model.
As can be seen from the Table \ref{tab:synapse}, our DSVM-UNet outperforms VM-UNetV2 by 0.47\% in terms of DSC and decreases it by 1.32\% in terms of HD95.
Besides, our approaches achieve the best performance across five categories of organ segmentation.
\vspace{-2mm}
\subsection{Ablation Studies}
To validate the effectiveness of each self-distillation method, we conducted ablation experiments on the ISIC18 dataset.
We first exploit $\mathcal{L}_{\text{BceDice}}$ as the baseline, which only achieves scores of 81.31\%, 89.63\%, 94.76\%, 96.11\%, and 89.65\% across the five evaluation metrics.
When we apply projective self-distillation $\mathcal{L}_{\text{Proj}}$ to it, the model can achieve an average improvement of 0.5\%. %
When we add another progressive self-distillation $\mathcal{L}_{\text{Prog}}$ to it, the model performance can be improved by an average of 0.75\%. 
Finally, when we apply both self-distillation to the model, it can further improve the performance of the model by an average of 0.97\% to reach the final SOTA results, which proves the effectiveness of our proposed methods. %
\begin{table}[htbp]
    \centering
    \vspace{-4mm}
    \caption{Ablation studies of proposed approaches on the ISIC18 dataset.}
    \vspace{2mm}
       \resizebox{0.95\linewidth}{!}{
            \begin{tabular}{ccccccccc}
           \toprule
              $\mathcal{L}_{\mathrm{BceDice}}$ &  $\mathcal{L}_{\mathrm{Proj}}$ &  $\mathcal{L}_{\mathrm{Prog}}$ & MIOU  & F1    & ACC   & SPEC  & SENS & Ave.\\
           \midrule
          $\checkmark$ & $\times$ &  $\times$  & 81.31 & 89.63 & 94.76 & 96.11 & 89.65 &  90.29 \\
          $\checkmark$ & $\checkmark$ &  $\times$ & 81.47 & 89.96 & 95.22 & 97.41 & 89.87 & 90.79 \\
          $\checkmark$ & $\times$ & $\checkmark$ & 81.49 & 90.21  & 95.38 & 97.60 & 90.54 & 91.04 \\
          $\checkmark$ & $\checkmark$ & $\checkmark$ & \textbf{81.51} & \textbf{90.45} &\textbf{95.43} & \textbf{97.66} & \textbf{91.26} & \textbf{91.26} \\
           \bottomrule
              \end{tabular}%
            }
        \label{tab:ablation}%
         \vspace{-6mm}
\end{table}%

\subsection{Model Complexity}
To further evaluate model complexity, we conducted comparative studies with existing methods, as shown in Table \ref{tab:complex}. We observed that VM-UNet has a relatively high parameter count of 27.42 M, but it maintains a low computational overhead of 4.11 G. Although VM-UNetv2 has only 22.77 M parameters, its more complex design leads to an increase in computational overhead to 4.4 G. In contrast, our approach achieves a balanced trade-off between performance and computational cost, demonstrating its superiority.
\begin{table}[htbp]
  \centering
  \vspace{-4mm}
  \caption{Computational complexity analysis of different methods on ISIC17 dataset.}
  \vspace{2mm}
  \resizebox{0.9\linewidth}{!}{
    \begin{tabular}{lcccc}
    \toprule
    Complexity & VM-UNet & VM-Unet++ & VM-UNetV2 & Ours \\
    \midrule
    \#FLOPs (G) & 4.11  & 4.33  & 4.40  & \textbf{3.65} \\
    \#Param. (M) & 27.42 & 24.81 & 22.77 & \textbf{22.63} \\
    \bottomrule
    \end{tabular}%
   
  }
  \label{tab:complex}%
   \vspace{-4mm}
\end{table}%

\section{Conclusion}
In this paper, we present DSVM-UNet, an effective approach that enhances the model by leveraging feature knowledge for dual self-distillation, all without relying on complex architectural designs. Each method we introduce contributes to improved performance to a certain degree. Our extensive experiments conducted on various benchmarks demonstrate the effectiveness of our proposed methods.
\bibliographystyle{IEEEbib}
\bibliography{refs}

\end{document}